\pdfoutput=1

\documentclass[11pt]{article}

\usepackage{amsmath}
\usepackage[final]{EMNLP2023}

\usepackage{times}
\usepackage{latexsym}
\usepackage{yhmath}
\usepackage{amsfonts}
\usepackage{amssymb}
\usepackage{bbm}
\usepackage[T1]{fontenc}

\usepackage[utf8]{inputenc}

\usepackage{microtype}

\usepackage{inconsolata}
\usepackage{microtype}
\usepackage{graphicx}
\usepackage{multirow}
\usepackage{booktabs}

\title{Neutralizing Bias in LLM Reasoning using Entailment Graphs}

\author{{\bf Liang Cheng}$^{\dag}$ \quad {\bf Tianyi Li}$^{\dag}$ \quad {\bf Zhaowei Wang}$^{\ddag}$ \quad {\bf Tianyang Liu}$^{\dag}$ \quad {\bf Mark Steedman}$^{\dag}$ \\
        $^{\dag}$University of Edinburgh \quad $^{\ddag}$HKUST \quad \\
        \texttt{\{L.Cheng-13,tianyi.li,t.liu-47\}@sms.ed.ac.uk \quad zwanggy@cse.ust.hk \quad steedman@inf.ed.ac.uk}}

\begin{document}
{\makeatletter\acl@finalcopytrue
  \maketitle
}
\begin{abstract}
LLMs are often claimed to be capable of Natural Language Inference (NLI), which is widely regarded as a cornerstone of more complex forms of reasoning. However, recent works show that LLMs still suffer from hallucinations in NLI due to \textit{attestation bias}, where LLMs overly rely on propositional memory to build shortcuts. To solve the issue, we design an unsupervised framework to construct counterfactual reasoning data and fine-tune LLMs to reduce attestation bias. To measure bias reduction, we build \textit{bias-adversarial} variants of NLI datasets with randomly replaced predicates in premises while keeping hypotheses unchanged. Extensive evaluations show that our framework can significantly reduce hallucinations from attestation bias. Then, we further evaluate LLMs fine-tuned with our framework on original NLI datasets and their bias-neutralized versions, where original entities are replaced with randomly sampled ones. Extensive results show that our framework consistently improves inferential performance on both original and bias-neutralized NLI datasets.

\end{abstract}

\section{Introduction}
\label{sec:introduction}
Natural Language Inference (NLI) has long been recognized as a foundational understanding task in language understanding with various downstream applications~\cite{cheng2023complementary, deng2023nonfactoid, gao2023retrieval}. It assesses the understanding ability of models by requiring them to determine if a given premise logically entails a hypothesis.  
Recently, with the rise of LLMs, the field of NLI has witnessed significant advancements \cite{brown2020language,he2023using,liu2024large}. These models, pre-trained on vast amounts of text data, have been claimed to capture inferential relations between statements enabling reasoning, positioning them as state-of-the-art tools for NLI. 

However, despite the apparent success of LLMs on natural NLI tasks, they continue to embody response biases, leading to the phenomenon of hallucination \cite{huang2023towards,ji2023survey,gallegos2024bias}. In NLI, this issue arises because models rely more on memorization from their training corpus rather than inference from the given premises, causing false positive entailment judgments when the hypothesis is \textit{attested} in the training data \cite{poliak2018hypothesis, Rawte2023ASO}, the phenomenon known as \textbf{attestation bias}~\cite{mckenna2023sources}. The attestation bias leads to brittleness in bias-adversarial cases, and poses challenges in accurately assess the bias-free reasoning capabilities of LLMs \cite{mckenna2023sources}. 

\begin{figure}[t]
	\centering
	\includegraphics[width=1.0\linewidth]{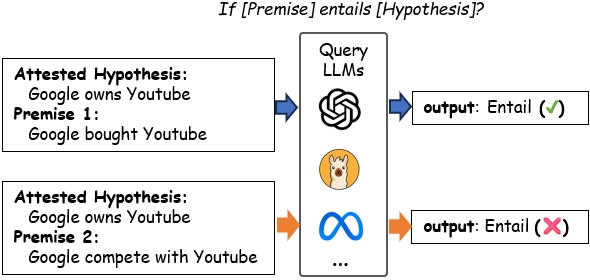}
	\caption{An example of attestation bias. LLMs tend to evaluate entailment with their memorized knowledge rather than given premise.}
	\label{Fig:example of Att Bias}
 \vspace{-0.1in}
\end{figure}

In this study, we explore the method of fine-tuning LLMs to improve their robustness against the attestation bias. We propose an \textit{unsupervised} approach to construct counterfactual but logically consistent datasets for training LLMs. 
Our approach begin with unsupervised extraction of textual entailment relations between predicates from large-scale open-domain corpora using semantic parsing. The extracted data is then formatted into Entailment Graphs (EGs) \cite{hosseini2018learning, hosseini2021open}, which consist of typed predicate pairs. Finally, we generate counterfactual samples by randomly selecting named entities and other arguments to instantiate these types.

\begin{figure*}[t]
	\centering
	\includegraphics[width=1.0\linewidth]{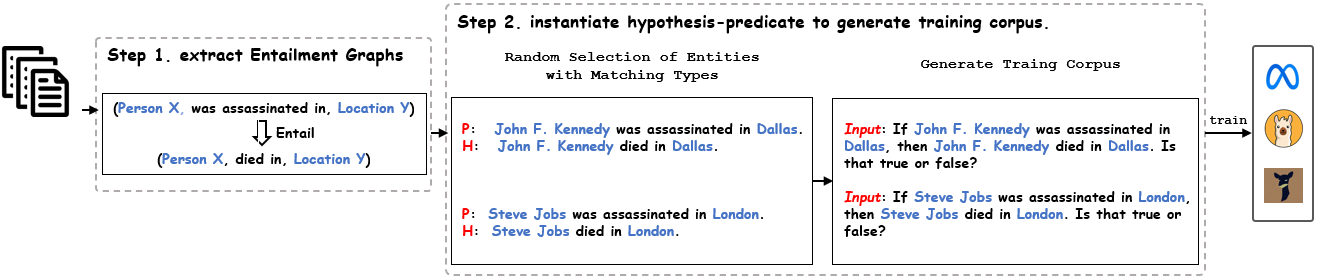}
	\caption{The pipeline of our approach: \textbf{Step 1:} Build EGs in unsupervised manner. \textbf{Step 2:} Instantiate predicates using random entities with matching types, then wrap instantiated predicates into prompts to generate training corpus. }
	\label{Fig:pipeline}
 \vspace{-0.1in}
\end{figure*}

We evaluate the effectiveness of our method along two dimensions: bias reduction and general inferential performance improvement.

First, to measure how well our training has reduced the attestation bias, we compare LLMs before and after our training on \textit{bias-adversarial} variants of NLI datasets.
Specifically, we randomly alter the predicates in the premises while keeping the hypothesis fixed.  The newly generated premises are non-entailing, so any positive judgments by the LLM are false positives, arising from attestation bias relating to the hypothesis. The results demonstrate that training LLMs with our method can significantly reduce attestation bias, ensuring a more reliable evaluation of their reasoning capabilities.

Second, to evaluate the effectiveness of our training in improving inferential performance on the original NLI tasks, we conduct experiments on both the \textit{standard} and a further \textit{bias-neutralized} NLI datasets, where entities in the original dataset are replaced with randomly selected entities of the same type. 
In both cases, our approach outperforms baseline models, demonstrating its robustness and effectiveness in enhancing inferential capability.

The main contributions of this paper are summarized as follows:

(a) To reduce attestation bias while enhancing inferential capabilities, we propose an unsupervised approach using EGs to generate a logically sound inference dataset, free of the artefacts that plague human-constructed NLI datasets, for fine-tuning LLMs.

(b) We show that for 4 different LLMs, our approach reduce attestation bias and improves performance in NLI tasks.

(c) We introduce a bias-neutralized method for a more accurate evaluation of LLMs' true inferential capabilities. This approach generates bias-neutralized test sets, where our EG-enhanced models consistently achieve superior reasoning performance on both standard and these bias-neutralized inference datasets.

\section{Related Work}
\label{related works}
\paragraph{Hallucination in Inference:} 
Hallucination in LLMs has emerged as a significant area of concern in NLP, as these models often generate content that is either factually inaccurate or contextually inappropriate. \citet{talman_testing_2019} report that many models struggle to generalize across different NLI datasets, even when the task format remains the same. In smaller language models, \citet{li_language_2022} observed a reliance on dataset artifacts when performing directional NLI on predicates. Furthermore, \citet{poliak2018hypothesis} found a range of NLI datasets containing artifacts that are memorized by supervised models trained on only sample hypotheses, causing overestimation of their inference performance. \citet{gulati2024putnamaxiom} proves that SOTA LLMs rely on memorized data to answer math questions, and their performance drops significantly once variable names are altered. \citet{carlini2023quantifying} found that LLMs are capable of memorizing significantly more data compared to smaller models, raising doubts on whether their performance gains stem from advanced inferential capabilities or more memorization.

\paragraph{Attestation Bias:} 
The attestation bias occurs when LLMs show a significantly higher probability of predicting \texttt{Entail} when the hypothesis is \textit{attested}, indicating that the inference process of LLMs is heavily influenced by their reliance on memorization about hypotheses. As a result, LLMs are inherently prone to disregarding the premise and responding incorrectly by relying on memorized information about hypothesis from their training corpus, as illustrated in Figure \ref{Fig:example of Att Bias}. 

\citet{mckenna2023sources} conducted a hypothesis-only test on LLMs, revealing that when labels contradict attestation bias, LLMs can be poor or even near-random classifier. Their research demonstrates that the attestation bias is the primary source of hallucination in LLMs on inference tasks. 

\paragraph{Entailment Graphs:} EGs are symbolic graphs used to preserve entailment relations between predicates \cite{berant2010global,berant2011global, hosseini2018learning, hosseini2021open}. 
Unlike sentence-level inference data, EGs are formatted as sets of triples, with each triple consisting of predicate pairs and typed arguments. For example, ``\textit{(Person.X, visited, Location.Y)} $\models$ \textit{(Person.X, went to, Location.Y)}''. EGs have been utilized in open-domain question answering and knowledge inference \cite{cheng2023complementary, wang2024absinstruct}.  

\section{Methodology}
\label{sec:Methods}
We propose an \textit{unsupervised} approach for constructing counterfactual reasoning datasets to fine-tune LLMs, enabling them to generalize beyond memorized knowledge and enhance inferential capabilities. 
\subsection{Constructing Counterfactual Reasoning Datasets}
\label{sec:experiments}
As demonstrated in Figure \ref{Fig:pipeline}, our counterfactual reasoning dataset is built in two key steps: unsupervised extraction of EGs (\S\ref{sec:Learn Entailment Graph}) and instantiating EG rules into NLI training sets (\S\ref{sec:hp pairs}).

\subsubsection{Extracting Entailment Graph}
\label{sec:Learn Entailment Graph}
We adopt the approach proposed by \citet{hosseini2018learning} for constructing EGs, which involves following main steps:

First, we employ a combinatory categorial grammar (CCG) parser \citep{steedman2000syntactic}, GraphParser \citep{reddy2014large}, to extract binary relations between a predicate and its arguments from sentences. Each argument, typically a noun, is then linked to its corresponding FreeBase IDs and entity type using the Named Entity Linking tool Aidalight \cite{nguyen2014aida}, formatting the extracted triple as predicates with typed arguments. 
Then, we compute the distributional similarity score\footnote{We use the Weeds similarity score \citep{weeds2003general} as the entailment score in our construction.} by calculating the co-occurrence of predicates associated with entities of the same \texttt{types}, assuming that predicates linked to the same entities refer to the same event or episode. We generate negative examples by replacing predicates with the same typed arguments. Notably, the entire process is \textit{unsupervised}. 

\textbf{Corpus:} Following \citet{hosseini2021open}, we use the multiple-source \textbf{NewsSpike} \cite{zhang2013harvesting} corpus to extract the EGs. NewsSpike was deliberately built to include different articles from different sources describing identical news events. From this dataset, we extract 5,500 positive and 5,500 negative samples.

\subsubsection{Instantiating Premise-Hypothesis Pairs}
\label{sec:hp pairs}
Entailment graphs consist of entailment rules, each rule in an EG involves a pair of predicates expecting two typed arguments. To instantiate EG rules into NLI data entries, we first replace the type arguments with specific named entities.
To ensure consistent entity replacements, we categorize entities from the open-domain corpus into 48 FIGER types \cite{ling-figer}, such as ``person'' or ``location'', aligning them with the types in Freebase \cite{bollacker-freebase}. We assign a default type ``thing'' in failure cases. Using the entity-to-type mapping, we randomly select entities that match the corresponding types to instantiate typed EGs into premise-hypothesis pairs. 

EG rules are predicate focused and applies to any context, so the instantiation process preserves the entailment relationships within typed EGs. Figure \ref{Fig:pipeline} illustrates this process with an example, where the entailment ``\textit{(Person X, was assassinated in, Location Y)} $\models$ \textit{(Person X, died in, Location Y)}'' is formatted into the following NLI format: ``\textbf{[Premise]}: Steve Jobs was assassinated in London. \textbf{[Hypothesis]}: Steve Jobs died in London''. Our constructed premise-hypothesis pairs preserve the logical inferential relationship but are counterfactual.

We adopt the prompt templates from previous studies \cite{schmitt_language_2021, mckenna2023sources}, formatting the premise-hypothesis pairs as a two-way answer choice: A) True, B) False. These generated natural questions are then used to query the models. The full list of concrete prompts can be found in Appendix \ref{sec:prompt template for fine-tuning appendix}. 

\subsection{Training LLMs with Instantiated EGs}
\label{sec:fine-tuning LLMs}
We use the counterfactual reasoning data generated from EGs to fine-tune DeepSeek-R1-Distill-Llama-8B \cite{deepseekai2025deepseekr1incentivizingreasoningcapability}, Mistral-7B, LLaMA-3-8B-instruct, and LLaMA-3-70B-instruct models. These models are widely recognized for their strong reasoning capabilities and have drawn significant interest from researchers. 

Following the \textit{llama-recipes} tools published by Meta\footnote{\href{https://github.com/meta-llama/llama-recipes\%7D}{https://github.com/meta-llama/llama-recipes}}, we fine-tune LLMs using LoRA  \cite{hu2022lora} within the PEFT (Parameter-Efficient Fine-Tuning) \cite{ding2023parameter} framework.  During fine-tuning, we set a fixed learning rate of 1e$^{-4}$ and train for 12 epochs. The LoRA rank is set to 8, with LoRA dropout rate of 0.05. 

\begin{table*}[h]
	\normalsize
	\centering
	\begin{center}\resizebox{0.9\textwidth}{!}{%
        \begin{tabular}{ccc}
\hline
\textbf{Task}  & \textbf{Sample Query: {[}premise{]} $\Rightarrow$ {[}hypothesis{]}}                        & \textbf{Dataset Label}              \\ \hline
$I$            & George Bush \textcolor{blue}{was the governor of} Texas $\Rightarrow$ George Bush \textcolor{blue}{is a politician from} Texas& \texttt{Entail}\\
\textit{RPI} & George Bush \textcolor{blue}{lives in} Texas $\Rightarrow$ George Bush \textcolor{blue}{is a politician from} Texas& \texttt{No-Entail}\\ \hline
\end{tabular}%
}
	\end{center}
    \caption{A sample of generate \textit{RPI} from original inference task $I$.}
	\label{Tab:Sample of data transform}
 \vspace{-0.1in}
\end{table*}

\section{Experimental Setup}
\label{sec:experiments}
To evaluate the effectiveness of our EG-enhanced models, we conduct two experiments: (1) examining whether EG-enhanced models reduce attestation bias in §\ref{sec: Reduce attestation bias}, and (2) assessing their inferential performance in NLI benchmarks in §\ref{sec: Evaluate Capability of Inference}.
Both evaluation experiments follow the same settings, including the evaluation datasets and baselines, as detailed in the following sections\footnote{All code and data are provided in \href{https://github.com/LeonChengg/EGs-tuned-LLMs.git}{https://github.com/LeonChengg/EGs-tuned-LLMs.git}}.

\subsection{Evaluation Datasets}
\label{sec:test set levy holt}
\paragraph{Levy/Holt} \cite{levy_annotating_2016, holt_probabilistic_2019} dataset is a widely used for NLI, which comprises premise-hypothesis pairs structured in a specific task format: ``Given [premise $P$], is it true that [hypothesis $H$]?''. Each $P$- and $H$-statement has the property of containing one predicate with two named entity arguments, where the same entities appear in both $P$ and $H$. The Levy/Holt dataset contains inverse of all entailment pairs. Following \citet{mckenna2023sources}, we study the challenging \textit{directional} subset, where the entailments hold in one direction but not both. 
\paragraph{SNLI} \citep{bowman-etal-2015-large} is another widely used datasets for NLI. It consists of human-generated premise-hypothesis pairs with manually assigned labels. Unlike Levy/Holt, which contains named-entity artifacts, SNLI is composed of general sentences in both premises and hypotheses, typically without named entities.  

\paragraph{Fomatting as two-choice questions:} For evaluation, premise-hypothesis pairs in Levy/Holt and SNLI are formatted as two-choice natural questions using prompts, where choice A corresponds to \texttt{Entail} and choice B to \texttt{No-Entail}, ensuring alignment with the Levy/Holt and SNLI annotations\footnote{In SNLI, contradiction and neutral labels are categorized as No-Entail.}. These prompts are crafted by human experts \cite{schmitt_language_2021,mckenna2023sources}, as shown in Appendix \S\ref{sec:prompt template for attesting inference}. During evaluation, all models successfully selected either A or B for every development set question, indicating compatibility with the QA format.

\subsection{Baselines}
\paragraph{Standard LLMs with Few-Shot Setting:}
To evaluate the performance of EG-enhanced LLMs, we compare them against the original LLMs as baseline models. 
During inference, we adopt a few-shot approach, hand-annotating a minimal set of 4 examples in the style of the template. 
 These examples are prepended before the query (see Appendix \ref{sec:prompt template for attesting inference} for an example).
Our goal is to analyze model behavior as conditions change, rather than maximize the score on a specific dataset. Therefore, we maintain a minimal 4-example setup to evoke positive responses across LLMs. 

\paragraph{Chain-of-Thought Reasoning:} In our experiments, we incorporate manually written explanations in few-shot examples, providing step-by-step reasoning before each answer to guide LLMs. Specifically, we utilize a three-step analytical process for guidance: analyzing the premise, analyzing the hypothesis and clarifying the relationship between premise and hypothesis. 
These explanations, detailed in Appendix \ref{sec:CoT prompt}, serve as Chain-of-Thought (CoT) prompts, establishing a baseline for evaluating LLM performance under CoT guidance.

\section{Experiment 1: \newline Attestation Bias Reduction}
\label{sec: Reduce attestation bias}
To evaluate the effectiveness of our method in reducing attestation bias, we measure attestation bias by comparing estimated probabilities of predicting \texttt{Entail} conditioned on whether the hypothesis is predicted \texttt{Attested} or not. 

However, in original NLI dataset entailments may coincidentally refer to attested facts, which could lead to spurious correlation between inference and attestation scores, making it difficult to determine whether LLMs rely on memory or reasoning to generate predictions.
To address this issue, we adopt the \textbf{Random Premise Inference Task} proposed by \citet{mckenna2023sources}, which serves as a \textit{bias-adversarial}  benchmark for accurately quantifying attestation bias.

\textbf{The Random Premise Task (\textit{RPI})} modifies the original NLI dataset by replacing the original premise predicate with a randomly selected predicate while keeping the hypothesis fixed and maintaining the same entity arguments.
As a result, this process create non-entailing premise-hypothesis pairs, as the example illustrated in Table \ref{Tab:Sample of data transform}. This transformation produces a dataset in which all samples are labeled as negatives (\texttt{No-Entail}), as two randomly paired predicates are highly unlikely to form entailment relations\footnote{\citet{mckenna2023sources} manually inspect the generated random premise entries for the Levy/Holt dataset and found only 9.6\% to be true entailment.}. We determine whether LLMs can correctly identify these samples as negatives or if they rely solely on attested hypotheses to make false positive predictions.  An ideal model should predict zero \texttt{Entail}. 
The \textit{RPI} task effectively tests the model's reliance on propositional memory, as it prevents true entailments while maintaining the attestedness of the conclusions (hypotheses).

To calculate attestation biases, we first determine the attestedness of each hypothesis by prompting the LLM to classify it as true, false, or unknown, following the same prompts\footnote{The attestation prompt is provided in Appendix  \ref{sec:prompt template for attesting hypothesis appendix}. } used in prior studies \cite{poliak2018hypothesis, mckenna2023sources}. Then, we categorize all samples into two groups: \textit{attested} and \textit{non-attested}, depending on whether the LLM identifies the hypothesis as true. We calculate the proportion of \texttt{Entail} predictions in the \textit{attested} set and compare it to the \textit{non-attested} set, providing an effective measure of attestation bias by highlighting differences in prediction behavior between the two sets. 

We adopt the Levy/Holt dataset for this attestation bias measurement experiment because this dataset includes artifacts containing named entities, allowing us to assess whether these artifacts in hypothesis are attested by LLMs. 

\begin{table}[t]
	\normalsize
	\centering	\begin{center}\resizebox{1.0\linewidth}{!}{%
        \begin{tabular}{ccc}
\hline
\textbf{Model}&  \textbf{AttBias} & \textbf{$\Delta_{AttBias}$}\\ \hline
 DeepSeek-R1-Llama-8B& 26.04&-\\
 DeepSeek-R1-Llama-8B$_{CoT}$& 15.10&-10.94\\
 DeepSeek-R1-Llama-8B$_{EG}$& \textbf{7.58}&\textbf{-18.46}\\ \hline
 Mistral-7B& 32.98&-\\
 Mistral-7B$_{CoT}$& 22.64&-10.34\\
 Mistral-7B$_{EG}$& \textbf{13.0}&\textbf{-19.98}\\ \hline
 Llama-3-8B                    & 23.12 &-\\
 Llama-3-8B$_{CoT}$& 13.94&-9.18\\
 Llama-3-8B$_{EG}$& \textbf{5.99}&\textbf{-17.13}\\ \hline
Llama-3-70B                   & 19.20 & -\\
 Llama-3-70B$_{CoT}$& 15.96&-3.24\\
Llama-3-70B$_{EG}$& \textbf{8.34}&\textbf{-10.86}\\ \hline
\end{tabular}%
}
	\end{center}
    \caption{AttBias scores on the \textit{RPI} Levy/Holt dataset across various models (lower is better). The subscript \textit{EG} denotes LLMs trained on our constructed EGs, while \textit{CoT} refers to models employing chain-of-thought reasoning.}
	\label{Tab:Result of random premise AttBias}
    \vspace{-0.1in}
\end{table}

\subsection{Scoring Attestation Bias}
\label{sec:Random Premise Task scoring}
Attestation bias reflects a significantly higher likelihood of predicting \texttt{Entail} for \textit{attested} hypotheses compared to \textit{non-attested} hypotheses. To quantify this bias, we define the \textbf{Attestation Bias score} as follows: 
\begin{align*}
    AttBias = P(tok=\texttt{Entail} | Att(hypo)) \\
    - P(tok=\texttt{Entail} | \neg{Att(hypo)})
\end{align*}
where \textit{P}(\textit{tok} = \texttt{Entail} | \textit{Att}(\textit{hypo})) represents the estimated conditional probability of predicting \texttt{Entail} when the hypothesis is attested. When the models achieve the same accuracy rate\footnote{In our experiments, accuracy rate is set to 0.5.}, we calculate the proportion of \texttt{Entail} predictions that fall within the \textit{attested} hypothesis set compared to \textit{non-attested} set.

Lower AttBias scores representing reduced influence from attested memorization, indicating less impact of attestation bias.

\subsection{Results of Experiment I: Degree of Bias Reduction}
\label{sec:Random Premise Task Results}
Table \ref{Tab:Result of random premise AttBias} presents the Attestation Bias scores for the \textit{RPI} task. The results demonstrate that our EG-enhanced LLMs significantly reduce attestation bias, addressing the challenges of hallucination in NLI. Additionally, we report the reduction of false positives on \textit{attested} hypotheses in  Appendix \ref{sec: false positives ratio}, indicating reduced reliance on the model’s memorization of the training corpus.

We observe that the reduction in attestation bias is more pronounced in smaller models, compared to larger models, such as LLaMA-3-70B. This suggests that the fine-tuning with our EGs is more effective for smaller-sized models, as larger models require more extensive EGs data for fine-tuning. To validate this, we fine-tune LLMs with increasing amounts of training data generated from EGs. Our results indicate that attestation bias in LLaMA-3-70B continues to decline as training data increases, reaching levels comparable to those observed in smaller LLMs (detailed in Appendix \ref{sec: Fine-tune LLMS with different sizes}). 

\begin{figure}[]
	\centering
	\includegraphics[width=1.0\linewidth]{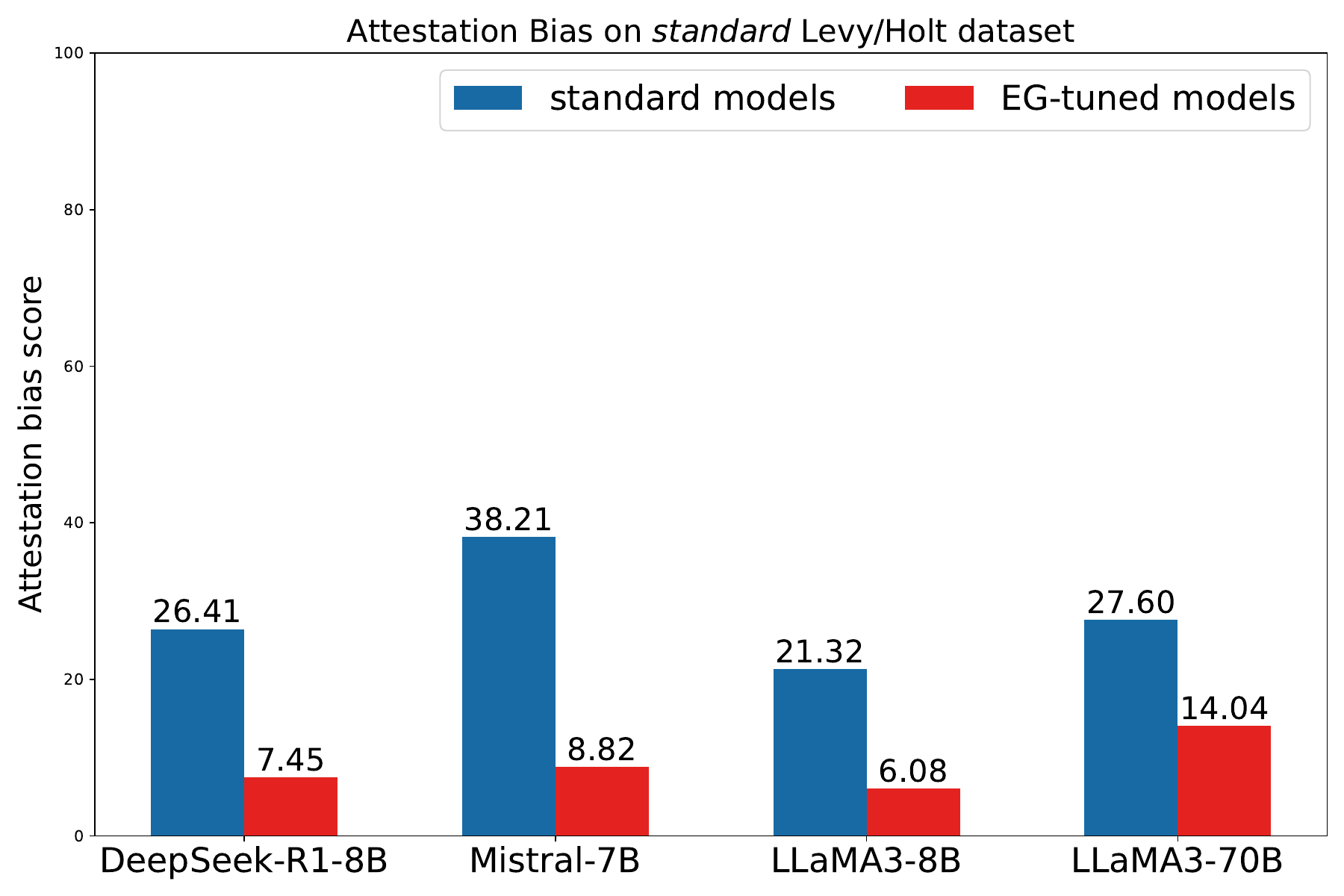}
\caption{The Attestation Bias scores for the original Levy/Holt, demonstrating a consistent attestation bias reduction after fine-tuning with EGs.}
    \vspace{-0.1in}
	\label{Fig:Result of AttBias on original LH}
\end{figure}

In addition to the \textit{RPI} task, which is explicitly designed to be bias-adversarial, we also present the AttBias scores on the original Levy/Holt dataset in Figure \ref{Fig:Result of AttBias on original LH}, which demonstrate a consistent reduction in attestation bias following fine-tuning with EGs.

The significant reduction in attestation bias can be attributed to the counterfactual nature of our EG-based training corpus. During fine-tuning, the LLMs learn entailment between predicates while incorporating counterfactual knowledge, thereby reducing their reliance on memorized artifacts for inference. Our method minimizes hallucinations from attestation bias and enhances overall robustness.

\section{Experiment 2: \newline Evaluate Capability of Inference}
\label{sec: Evaluate Capability of Inference}
EG-enhanced LLMs exhibit a notable reduction in attestation bias. To further investigate how this attestation bias reduction affects their capability of inference, we evaluate their general inferential performance on original NLI tasks. 

However, attestation biases and reported data leakage \citep{gururangan-etal-2020-dont,balloccu2024leak,ravaut2024much} often lead to an overestimation of LLMs performance on original NLI dataset, where LLMs rely on memorization to form reasoning shortcuts rather than genuinely reasoning, making it challenging to accurately evaluating their \textit{true} inference capabilities.

To address these challenges, we introduce a method to generate bias-neutralized dataset from original NLI datasets for evaluation. We neutralize biases by replacing the original entities with others of the same type, generating new inference data that includes counterfactual statements while preserving original entailment labels. 
We then evaluate our EG-enhanced LLMs on both \textit{\textbf{original}} and \textit{\textbf{bias-neutralized}} NLI test sets, enabling a more accurate assessment of their \textit{true} inferential capabilities. 

\begin{table*}[]
	\normalsize
	\centering
	\begin{center}\resizebox{0.9\textwidth}{!}{%
        \begin{tabular}{ccc}
\hline
\textbf{Task}  & \textbf{Sample Query: {[}premise{]} $\Rightarrow$ {[}hypothesis{]}}                        & \textbf{Dataset Label}              \\ \hline
\textit{LH}& \textbf{George Bush} was the Governor of \textbf{Texas} $\Rightarrow$ \textbf{George Bush} is a politician from \textbf{Texas}& \texttt{Entail}\\ \hline
\textit{LH}$_{rpArg\downarrow}$& \textbf{Jan Hus} was the Governor of \textbf{Svaneti}  $\Rightarrow$ \textbf{Jan Hus} is a politician from \textbf{Svaneti}& \texttt{Entail}\\
\textit{LH}$_{rpArg\uparrow}$& \textbf{Elon Musk} was the Governor of \textbf{Paris} $\Rightarrow$ \textbf{Elon Musk} is a politician from \textbf{Paris}& \texttt{Entail}\\ \hline
\end{tabular}%
}
	\end{center}
    \caption{An exapmle of generating \textit{LH}$_{rpArg\downarrow}$ and \textit{LH}$_{rpArg\uparrow}$ from original Levy/Holt data (\textit{LH}).}
	\label{Tab:Sample of data transform rpArg}
 \vspace{-0.1in}
\end{table*}

\subsection{Neutralizing Biases}
\label{sec: Neutralize Biases}
To neutralize the biases arising from memorization, we generate counterfactual premise-hypothesis pairs from the original dataset by randomly replacing the original entities with other entities of the same type. Since the entities are randomly selected, the newly generated pairs are logically sound but are likely not to exist in original training corpora. This method generates novel  statements without changing their entailment labels, enabling an unbiased assessment of the model’s true inferential capability without interference of memorization.

We use entity type constraints here to ensure polysemous predicates maintain the same sense. For instance, the verb ``run'' has different meanings in ``[person] runs [organization]" versus ``[person] runs [software]'', but when substituting entities of the same type, the sense remains consistent. Therefore, the specific entity names do not affect the entailment labels \cite{Yarowsky:93a}.
Notably, this approach ensures that the dataset remains distinct from the original inference data and can allow for building new datasets for every experiment, thereby reducing potential overestimation caused by evaluation data leakage.

\subsection{Bias-neutralized Test Sets}
\label{sec: random argument inference}
We apply this method to construct bias-neutralized test sets from the original NLI datasets.

\paragraph{Replaced Argument LevyHolt} (\textit{LH}$_{rpArg}$):
All original entities in Levy/Holt are named entities. We replace them with other real entities of the same type, which are extracted using the named entity linking tool Aidalight \cite{nguyen2014aida} from NewsCrawl \cite{barrault_findings_2019}, a decade-long span of multi-source news text, in which entities are typed into FIGER types \cite{ling-figer}. 

Pre-trained LLMs are likely to contain more memorized knowledge about high-frequent named entities. Thus these generated counterfactual samples involving high-frequency entities have higher probability to conflict with the model's memorized knowledge, exacerbating attestation biases.
To further analyze the effect of entities frequency, we sample new entities uniform randomly from the 5\% least common entities in NewsCrawl ($LH_{rpArg\downarrow}$), and the 5\% most common ($LH_{rpArg\uparrow}$), separately. We insert the sampled entities while preserving the rest of each statement. Examples are shown in Table \ref{Tab:Sample of data transform rpArg}.

\paragraph{Replaced Argument SNLI} (\textit{SNLI}$_{rpArg}$):
Since the SNLI dataset typically consists of general sentences without named entities, making it impractical to directly extract real named entities from NewsCrawl to replace these general entities.

To address this, we adopt the approach proposed by \citet{liu2024explicit}, using ChatGPT to identify entities with their types that co-occur in both the hypothesis and premise (these are general entities, not limited to named entities) and then generate new entities that match the specified types to replace them. 
For instance, an original SNLI sample such as ``\textit{John gives Mary an \underline{apple} $\Rightarrow$ Mary receives an \underline{apple} from John}'', will be modified to ``\textit{John gives Mary a \underline{book} $\Rightarrow$ Mary receives a \underline{book} from John}''. 

We manually check 50 samples from \textit{LH}$_{rpArg}$ and \textit{SNLI}$_{rpArg}$ separately, confirming that all of them are logically sound. Additionally, we also examine the attestedness of hypotheses in \textit{LH}$_{rpArg}$ and \textit{SNLI}$_{rpArg}$ and find only 0.89\% are attested, ensuring that these bias-neutralized test sets are logically consistent and bias-free.

\subsection{Scoring Inferential Capability}
\label{sec:Inferential capability scoring}
Following \citet{{mckenna2023sources}}, we analyze model performance across varying confidence thresholds by converting letter choices into probabilities using the following mapping:
\begin{align*}
    S_{\text{ent}} = 0.5 &+ 0.5*\mathbb{I}[\text{tok}=\textbf{A}] * S_{\text{tok}} \\
    &- 0.5 * \mathbb{I}[\text{tok} = \textbf{B}] * S_{\text{tok}}
\end{align*}
Where $\mathbb{I}$ is the indicator function, and $S_{ent}$ estimates the probability of \texttt{Entail} from a textual output ($0 \leq S_{\text{ent}} \leq 1$) with token probability $S_{tok}$. The linear transformation preserves the ordering of model confidences, which is sufficient for calculating a precision-recall curve and \textbf{Area Under the Curve (AUC)} score.

\begin{table}[t]
    \normalsize
	\centering
	\resizebox{0.9\linewidth}{!}{
        \begin{tabular}{ccl}
\hline
Models& Levy/Holt&SNLI
\\ \hline
 DeepSeek-R1-Llama-8B& 69.25&85.06\\
 DeepSeek-R1-Llama-8B$_{CoT}$& 65.65&85.26\\
 DeepSeek-R1-Llama-8B$_{EG}$& \textbf{71.49}&\textbf{85.80}\\ \hline
Mistral-7B& 69.78
 &85.47\\
 Mistral-7B$_{CoT}$&65.14 &83.63\\
Mistral-7B$_{EG}$& \textbf{72.82}
 &\textbf{85.64}\\ \hline
 LLaMA-3-8B           &66.87 
 &\textbf{87.49}\\
 LLaMA-3-8B$_{CoT}$&62.40 &85.63\\
 LLaMA-3-8B$_{EG}$&\textbf{73.69} 
 &86.62\\ \hline
LLaMA-3-70B          & \multicolumn{1}{c}{77.40} & \textbf{90.01}
\\
 LLaMA-3-70B$_{CoT}$&76.53 &89.03\\
LLaMA-3-70B$_{EG}$& \textbf{77.46}  &89.85\\ \hline
\end{tabular}}
    \caption{AUC scores of original, EG-enhanced ($_{EG}$) and with chain-of-thought prompts ($_{CoT}$) LLMs versions on original Levy/Holt and SNLI.}
	\label{Tab:AUC original LH}
 \vspace{-0.1in}
\end{table}

\begin{table*}[h]
	\normalsize
	\centering
	\resizebox{0.7\linewidth}{!}{\begin{tabular}{ccccccc}
\hline
Models& \multicolumn{6}{c}{Tasks}   \\ \hline
                    & \multicolumn{2}{c|}{\textit{LH}$_{rpArg\downarrow}$}  & \multicolumn{2}{c|}{\textit{LH}$_{rpArg\uparrow}$}  & \multicolumn{2}{c}{\textit{SNLI}$_{rpArg}$}     \\ \cline{2-7} 
                    & AUC& \multicolumn{1}{c|}{$\Delta_{AUC}$} & AUC& \multicolumn{1}{c|}{$\Delta_{AUC}$} & AUC& $\Delta_{AUC}$\\ \hline
 DeepSeek-R1-Llama-8B& 66.17& -& 61.04& \multicolumn{1}{c|}{-} & 76.13&-\\
 DeepSeek-R1-Llama-8B$_{CoT}$& 59.57& -6.6& 56.65& \multicolumn{1}{c|}{-4.39} & 69.66&-6.47\\
 DeepSeek-R1-Llama-8B$_{EG}$& \textbf{67.37}& +1.2& \textbf{68.92}& \multicolumn{1}{c|}{+7.87} & \textbf{77.87}&+1.74\\ \hline
LLaMA-3-8B                & 61.80   & -& 59.05          & \multicolumn{1}{c|}{-} & 78.31 & -\\
 LLaMA-3-8B$_{CoT}$& 53.69& -8.11 & 54.95& \multicolumn{1}{c|}{-4.1} & 70.11&-8.20\\
LLaMA-3-8B$_{EG}$     & \textbf{71.27} &                               +9.47& \textbf{70.96} &  \multicolumn{1}{c|}{+11.91}       & \textbf{80.03 }&     +1.72\\ \hline
 Mistral-7B & 61.27& -& 59.96 & \multicolumn{1}{c|}{-} & 78.17& -\\
 Mistral-7B$_{CoT}$& 59.78& -1.49 & 57.52& \multicolumn{1}{c|}{-2.44} & 75.62&-2.55\\
 Mistral-7B$_{EG}$ & \textbf{71.20}& +9.93& \textbf{72.27}& \multicolumn{1}{c|}{+12.31} & \textbf{80.43}&+2.26\\\hline
LLaMA-3-70B        & 71.99          & -& 69.55   & \multicolumn{1}{c|}{-} & 80.26 & -\\
 LLaMA-3-70B$_{CoT}$& 70.65 & -1.34& 67.34& \multicolumn{1}{c|}{-2.21} & 80.10&-0.16\\
LLaMA-3-70B$_{EG}$    & \textbf{76.14} &                               +4.15& \textbf{76.71} &                        \multicolumn{1}{c|}{+7.16}        & \textbf{83.29} &                        +3.03\\ \hline
\end{tabular}}
    \caption{AUC scores  of LLMs, LLMs with chain-of-thought prompt ($_{CoT}$) and their EG-enhanced versions ($_{EG}$) on bias-neutralized inference datasets. }
	\label{Tab:Result of random Argument}
\end{table*}
\subsection{Results of Experiment II: Performance in Inference Tasks}
\label{sec:NLI result}
Table~\ref{Tab:AUC original LH} presents the AUC scores for original NLI datasets, demonstrating that EG-enhanced LLMs consistently outperform original LLMs across various model families on the Levy/Holt dataset. Notably, smaller models exhibit significant performance gains from EGs tuning, surpassing the improvements observed in larger models. On the original SNLI dataset, our EG-enhanced models yield only modest improvements.
One possible explanation for this limited improvement is data leakage \citep{balloccu2024leak}, as the original SNLI dataset may contain more memorized instances from pre-trained LLMs, leading to an overestimation of inferential capacities.

To assess the true capability of inference, we evaluate bias-neutralized inference test sets and present the AUC scores in Table~\ref{Tab:Result of random Argument}. The results highlight consistent improvements achieved by our EG-enhanced models across all bias-neutralized datasets. Additionally, the Precision-Recall curve in Appendix~\ref{sec:appendix_pr_curve} further illustrates that EG-enhanced LLMs outperform the original models, with particularly notable gains in smaller LLMs. 
Notably, across all bias-neutralized inference datasets, the EG-enhanced smaller models (DeepSeek-8B$_{EG}$, LLaMA-3-8B$_{EG}$ and Mistral-7B$_{EG}$) achieve performance on the same level of LLaMA-3-70B. These results suggest that after training on our EGs, smaller LLMs achieve inferential capabilities comparable to the standard extreme large models. We further examine the impact of prompt templates in Appendix \ref{sec:appendix-randEGs}, proving that LLMs are genuinely learning textual entailment during fine-tuning rather than merely memorizing the prompts.

In Table \ref{Tab:Result of random Argument}, we also observe that standard LLMs are limited in  \textit{LH}$_{rpArg\uparrow}$ and the improvement achieved by our method on \textit{LH}$_{rpArg\uparrow}$  is more pronounced compared to \textit{LH}$_{rpArg\downarrow}$. This suggests that standard LLMs face greater challenges when processing counterfactual information involving high-frequency entities. 
The observation indicates that LLMs overly rely on memorization to build reasoning shortcuts, ultimately undermining their reasoning performance on high-frequent counterfactual tasks, which have higher probability to conflict with the model's memorized knowledge. 
On the other hand, the improvement of our EG-enhanced models on \textit{LH}$_{rpArg\uparrow}$ highlights their enhanced inferential capability and robustness, particularly in handling conflicts between given premise and memorization.

\begin{figure}[]
	\centering
	\includegraphics[width=1.0\linewidth]{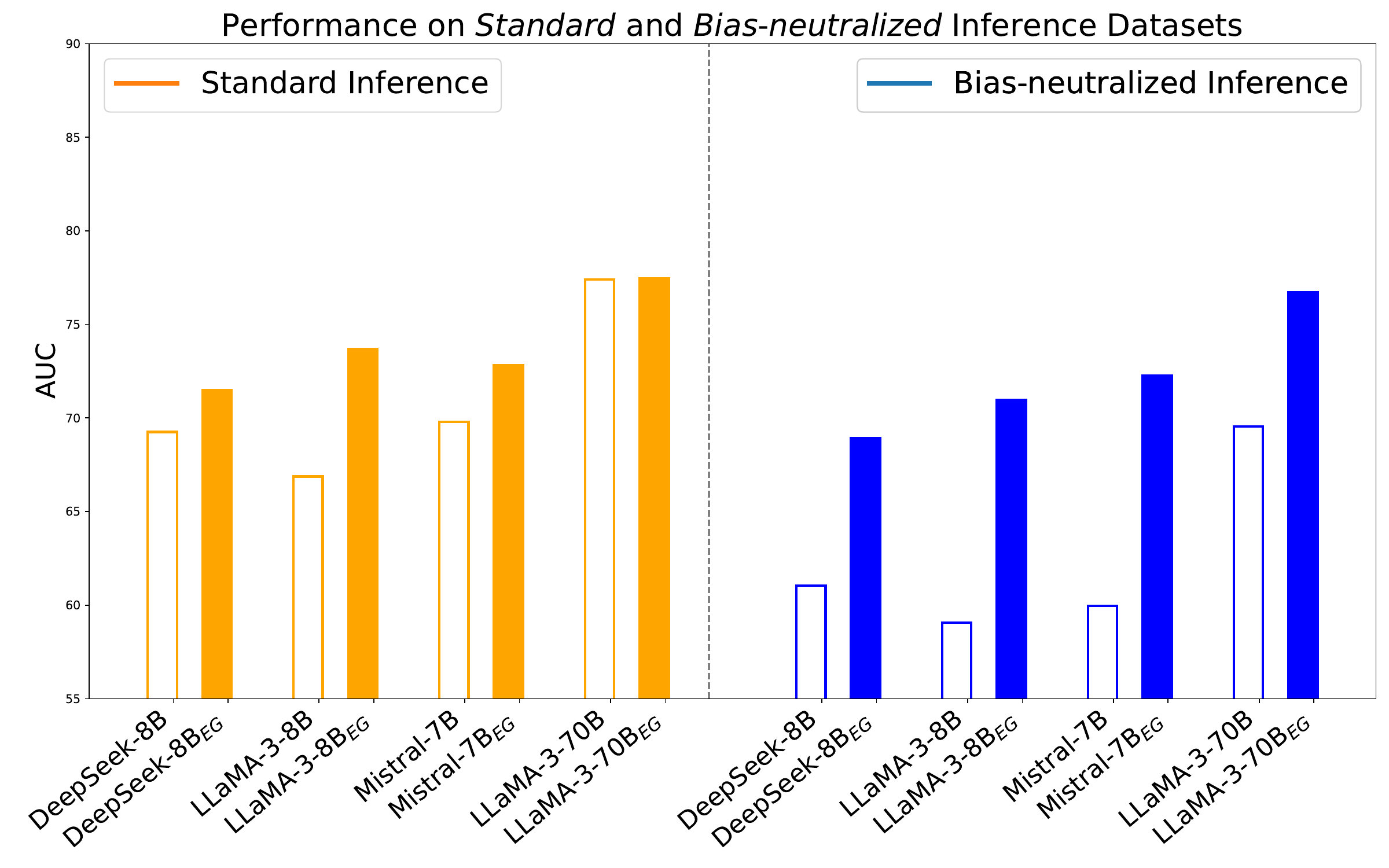}
	\caption{AUC scores of baseline (outline) and EG-trained (solid) LLMs on original (\textcolor{orange}{orange}) and our bias-neutralized (\textcolor{blue}{blue}) Levy/Holt.}
    \vspace{-0.1in}
	\label{Fig:Result of IM balance}
\end{figure}

Figure \ref{Fig:Result of IM balance} presents that EG-enhanced model exhibit a more pronounced improvement on bias-neutralized test set compared to original test sets. The reduction of attestation bias explains theses findings, as EG-enhanced models increasingly rely on their enhanced inferential capabilities to process NLI tasks rather than retrieving memorized information from training corpus. 

\section{Conclusion}
\label{sec:conclusion}
In NLI tasks, attestation bias in LLMs leads to false positives, as models rely on memorization rather than reasoning from the given premise, ultimately undermining their robustness and accuracy in inference.
To address this, we propose an unsupervised method for constructing logically consistent counterfactual EGs to fine-tune LLMs, enhancing their robustness against attestation bias. 
Experimental results show that our method reduces attestation bias and enhances inference by learning counterfactual EGs, enabling models to learn predicate entailment while without introducing artifacts, thereby minimizing dependency on memorized patterns. 
Our method improves robustness and effectiveness of models in practical applications while providing a more objective evaluation of inference capability across LLMs. 

\section{Limitation}
\label{sec:limitation}
In this paper, we propose an unsupervised approach to construct counterfactual reasoning data by instantiating entailment graphs and demonstrate its effectiveness in reducing hallucinations, enhancing the reasoning capabilities of LLMs.

A limitation of our current work is that the counterfactual reasoning data is used to fine-tune large LLMs for a specific task, natural language inference, rather than across a broader range of tasks. Although inference is foundational to many NLP tasks, it remains uncertain whether our approach will generalize effectively to other tasks. In future work, we plan to integrate counterfactual reasoning data into instruction-tuning frameworks to evaluate its performance across a wider variety of NLP tasks.

\clearpage
\bibliography{custom}
\bibliographystyle{acl_natbib}

\appendix
\clearpage

\section{Analyzing the effects of prompt templates}
\label{sec:appendix-randEGs}
One concern is whether the observed improvements in LLMs performance stem from learning the predicate entailment in EGs or simply memorizing the structure of query prompt templates. To determine this, we conducted a series of controlled experiments where the predicates in the EGs were randomly shuffled while preserving the original prompt structure during fine-tuning. These controlled datasets contain incorrect entailment relations between predicates but maintain the same prompt template. For instance, a shuffled example might present: ``If Steve Jobs \textit{was assassinated in} London, then Steve Jobs \textit{was born in} London. Is that true or false?''. As shown in Table \ref{Tab:AUC original rand EG}, the results show a significant performance drop (almost random predict) after fine-tuning on the randomly shuffled EGs. This confirms that the LLMs are indeed learning textual entailment during fine-tuning, rather than simply memorizing prompt templates.

\begin{table}[h]
	\normalsize
	\centering
	\resizebox{0.96\linewidth}{!}{
        \begin{tabular}{ccl}
\hline
Models& Levy/Holt&SNLI
\\ \hline
 DeepSeek-R1-Llama-8B& 69.25&85.06\\
 DeepSeek-R1-Llama-8B$_{EG}$& \textbf{71.49}&\textbf{85.80}\\
 DeepSeek-R1-Llama-8B$_{randEG}$& 52.0&51.76\\ \hline
Mistral-7B& 69.78
 &85.47\\
Mistral-7B$_{EG}$& \textbf{72.82}
 &\textbf{85.64}\\
 Mistral-7B$_{randEG}$&52.77 &50.34\\ \hline
 LLaMA-3-8B           &66.87 
 &\textbf{87.49}\\
 LLaMA-3-8B$_{EG}$&\textbf{73.69} 
 &86.62\\
 LLaMA-3-8B$_{randEG}$&51.03  &49.07\\ \hline
LLaMA-3-70B          & \multicolumn{1}{c}{77.40} & \textbf{90.01}
\\
LLaMA-3-70B$_{EG}$& \textbf{77.46}  &89.85\\ 
 LLaMA-3-70B$_{randEG}$&56.25  &52.91\\ \hline
\end{tabular}}
    \caption{The AUC score of original, EG-enhanced ($_{EG}$), with chain-of-thought prompts ($_{CoT}$) and the random-shuffled-EG-enhanced ($_{randEG}$) LLMs versions on original Levy/Holt and SNLI dataset.}
	\label{Tab:AUC original rand EG}
 \vspace{-0.1in}
\end{table}

\section{Precision-Recall curve}
\label{sec:appendix_pr_curve}
\begin{figure*}[t]
	\centering
	\includegraphics[width=0.75\linewidth]{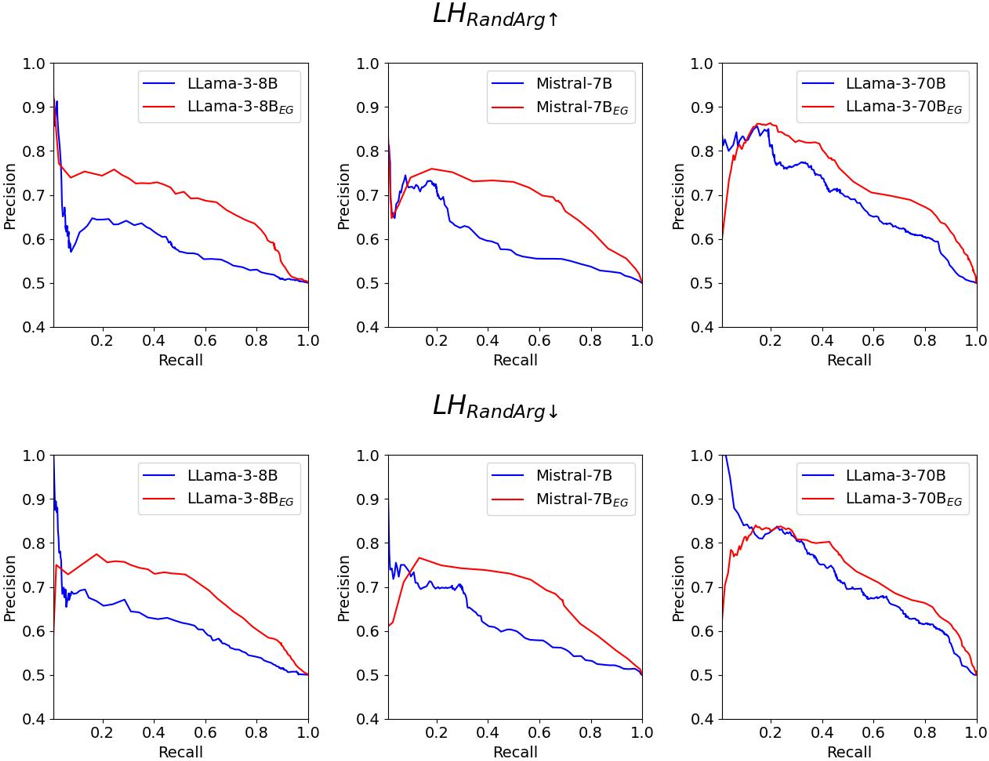}
	\caption{The Precision-Recall curve of our method compared to standard LLMs on \textit{LH}$_{rpArg\downarrow}$ and \textit{LH}$_{rpArg\uparrow}$ datasets.}
	\label{Fig:Result of random Argument}
\end{figure*}

\begin{figure}[h]
	\centering
	\includegraphics[width=1.0\linewidth]{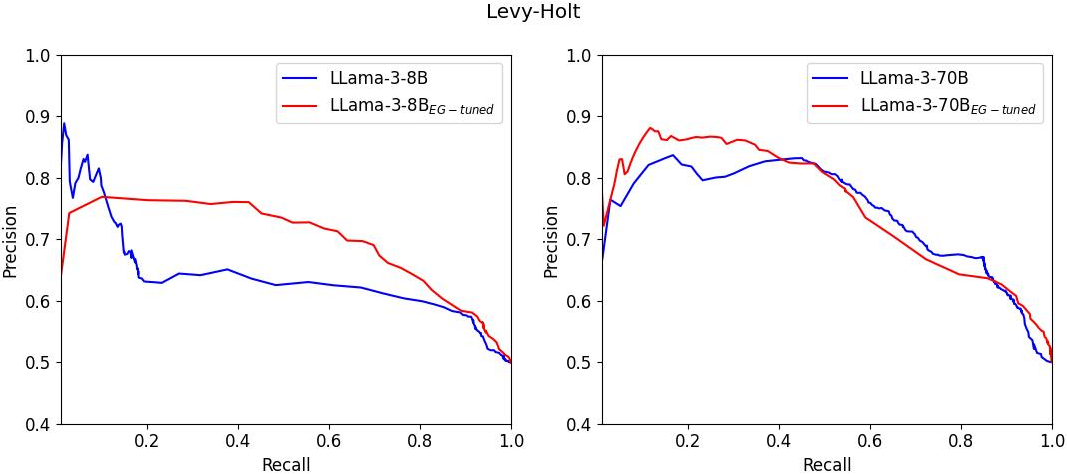}
	\caption{The Precision-Recall Curve of Levy/Holt across LLMs and the EG-tuned version.}
	\label{Fig:LH P-R result}
\end{figure}

To reduce biases arising from memorization, we evaluate bias-neutralized inference dataset and present the Precision-Recall curve for \textit{LH}$_{rpArg}$ in Figure~\ref{Fig:Result of random Argument}. The results show that our EG-enhanced LLMs outperform original models, with particularly significant improvements observed in smaller LLMs. Additionally, Figure~\ref{Fig:LH P-R result} shows the Precision-Recall curve for the original LH dataset, further demonstrating the consistent benefits of fine-tuning with EGs.

\begin{figure}[h]
	\centering
	\includegraphics[width=1.0\linewidth]{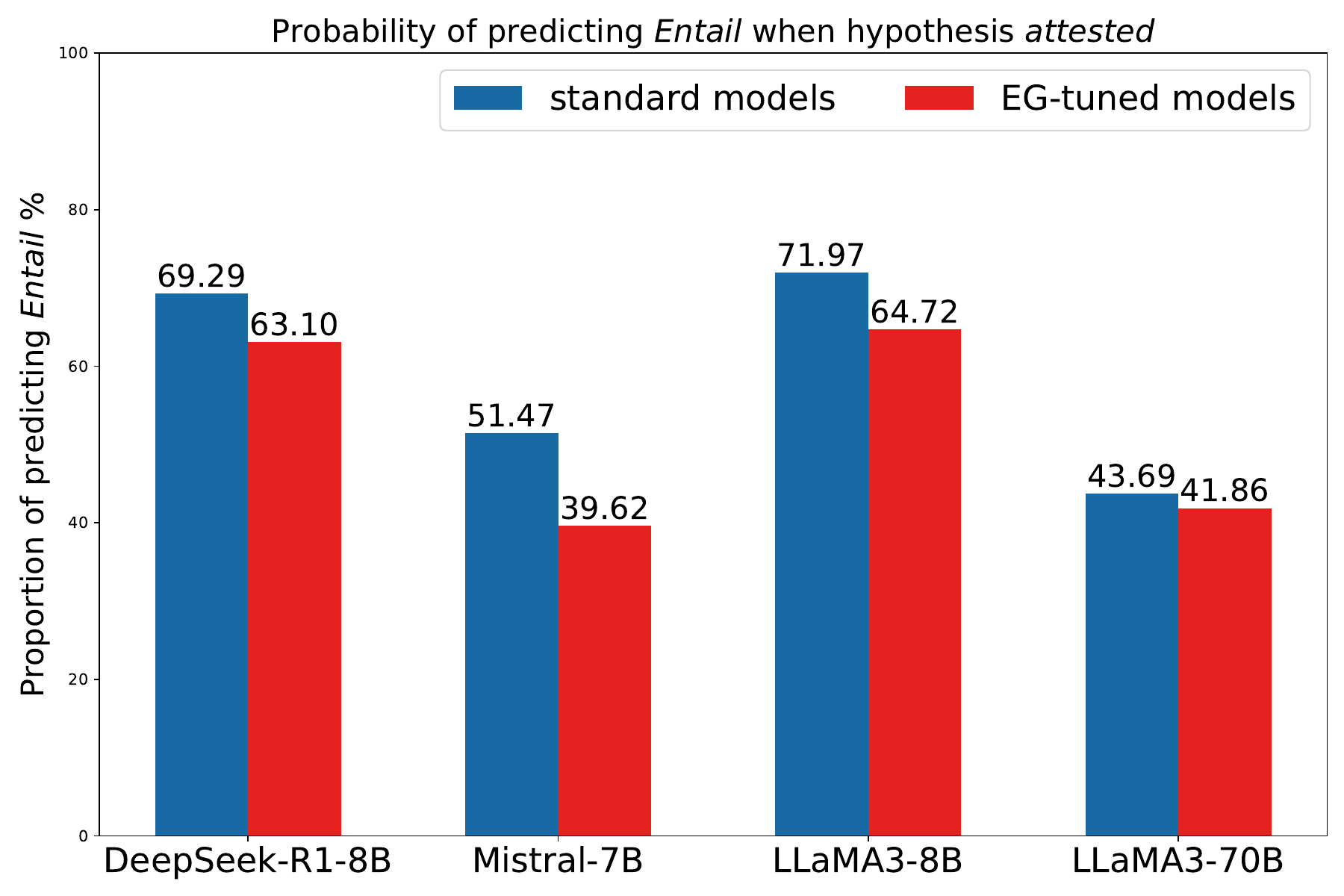}
	\caption{The probability of predicting \texttt{Entail} for \textit{RPI} LevyHolt, conditioned on the LLMs' attestation of the hypothesis. Since predicting \texttt{Entail} in this context represents a false positive hallucination, a lower probability is better. The image clearly shows that hallucination decrease significantly after fine-tuning with EGs. }
	\label{Fig:Result of random premise AttBias}
\end{figure}

\section{Proportion of False Positives on \textit{PRI}}
\label{sec: false positives ratio}

Figure \ref{Fig:Result of random premise AttBias} presents the estimated probability of predicting \textit{Entail} when the hypothesis is attested, highlighting a consistent decrease after fine-tuning with EGs across all LLMs. This decrease presents fewer false positives on \textit{attested} hypotheses, indicating reduced reliance on the model’s memorization of the training corpus.

\section{Fine-tuning LLMs with different sizes EGs }
\label{sec: Fine-tune LLMS with different sizes}
To evaluate the performance of LLMs trained on varying sizes of generated counterfactual data, we present the attestation bias scores in Table \ref{Tab:Result of AttBias on different sizes.}. The results indicate that smaller LLMs are more sensitive to fine-tuning, while extremely large LLMs require larger amounts of generated data during the fine-tuning process.

\begin{table}[]
	\normalsize
	\centering	\begin{center}\resizebox{0.95\linewidth}{!}{%
        \begin{tabular}{ccc}
\hline
\textbf{Sizes of EGs} & \multicolumn{2}{c}{\textbf{Model}} \\ \hline
                      & LLaMA-3-8B$_{EG}$& LLaMA-3-70B$_{EG}$\\ \cline{2-3} 
0 samples& 23.12           & 19.20            \\
2k samples& 14.03           & 18.63\\
5k samples& 12.30           & 11.93            \\
11k samples& 5.99            & 8.34             \\
18k samples& 4.87            & 5.81\\ \hline
\end{tabular}%
}
	\end{center}
    \caption{The AttBias scores of various LLMs fine-tuned with different sizes of EG datasets.}
	\label{Tab:Result of AttBias on different sizes.}
    \vspace{-0.1in}
\end{table}

\section{Prompt Format Selection}
\label{sec:appendix-prompt-format}

\paragraph{Prompt templates} are widely acknowledged for their significant and sometimes decisive impact on the behavior of LLMs. In our experiments, we categorize the prompt templates into three distinct types based on their usage: (a) prompt templates for fine-tuning LLMs, (b) prompt templates for hypothesis attestation, and (c) prompt templates used during inference. 

\subsection{prompt template for fine-tuning}
\label{sec:prompt template for fine-tuning appendix}
We adopt the manually crafted prompts utilized in prior inference studies \cite{schmitt_language_2021, mckenna2023sources},  which follow the format outlined below: 

\begin{enumerate}
    \item[] If $[$\textsc{premise}$]$, then $[$\textsc{hypothesis}$]$.
\end{enumerate}

 To make LLMs better understanding the task, we format it as Boolean questions and include indicator words such as ``Question:" and ``Answer:". For each choice, we automatically provide explanations for every answer by adding affirmation or negation to the propositions. As a result, the data in our counterfactual reasoning training corpus will be structured as Table \ref{Tab:prompt for training} shown. We fine-tune our method on these templates.

\subsection{prompt template for attesting hypothesis}
\label{sec:prompt template for attesting hypothesis appendix}
 In attesting hypothesis process, we using the same prompt discussed in \S\ref{sec:prompt template for fine-tuning appendix}, but mask the premise. Since the model may not be able to definitively determine whether the hypothesis is true or false, we use a question with three choices to evaluate the hypothesis, as (A) True, (B) Unknown, and (C) False.

In-context examples have been widely used for interacting with LLMs since \citet{brown2020language}. Moreover, \citet{NEURIPS2022_9d560961} demonstrated that incorporating chain-of-thought reasoning, step-by-step explanations, into in-context examples enhances LLMs' understanding of tasks.  
To improve the model's comprehension of the attestation task, we include a minimal set of three examples with explanations in few-shot settings, ensuring that each choice is represented by a corresponding example. We present the manual-crafted examples in Table \ref{tab:example_instantiated_prompts}.

\subsection{prompt template for inference}
\label{sec:prompt template for attesting inference}
Similar to the hypothesis attestation experiments (\S\ref{sec:prompt template for attesting hypothesis appendix}), we employ the few-shot settings in our prompts for inference tasks. Specifically, we manually created four examples for the inference tasks, consisting of two positive and two negative instances. These examples are presented in Table \ref{tab:example_instantiated_prompts}. 

\subsection{Chain-of-Thought Prompts}
\label{sec:CoT prompt}
Similar to the prompts used for inference (\S\ref{sec:prompt template for attesting inference}), we employ the few-shot settings in our prompts for inference tasks. Consistently, we manually created four examples for the inference tasks, each with step-by-step explanations to guide LLMs through reasoning. These examples are presented in Table \ref{tab:example_CoT_prompts}. 


\section{Computing Costs}
\label{sec:computing costs}
In our experiments, the extraction and learning of entailment graphs from the NewsSpike corpus required approximately 220G of CPU resources over a span of 20 hours. For the fine-tuning process, we employed four NVIDIA RTX A6000 GPUs to fine-tune the LLaMA-3-70B models, a process that took 21 hours to complete. This setup ensured efficient resource utilization while achieving optimal performance for our large-scale model fine-tuning. 

\begin{table*}[!t]
        \normalsize
	   \centering
\begin{tabular}{ll}
\hline
              & \begin{tabular}[c]{@{}l@{}}Question: If $[$\textsc{premise}$]$, then $[$\textsc{hypothesis}$]$. Is that true or false?\\ (A) True; (B) false\end{tabular} \\ \hline
label=\textit{True}  & \begin{tabular}[c]{@{}l@{}}(A) True.\\ Yes, it is true. $[$\textsc{premise}$]$ entails $[$\textsc{hypothesis}$]$.\end{tabular}                            \\ \hline
label=\textit{False} & \begin{tabular}[c]{@{}l@{}}(B) False.\\ No, it is false. $[$\textsc{premise}$]$ does not entail $[$\textsc{hypothesis}$]$.\end{tabular}                   \\ \hline
\end{tabular}
\caption{The table present the prompt template using in our training steps.}
	\label{Tab:prompt for training}
\end{table*}

\begin{table*}[t]
    \centering
    \small
    \begin{tabular}{p{12cm}}
        \toprule \hline
        \textbf{A. Few-shot Examples Instantiated Prompt for Inference Task}\\ \hline
        \midrule
        If Google bought Youtube, then Google owns Youtube. Is that true or false?\\
A) True\\
B) False\\
Answer: A) True. Owning is a consequence of buying.\\
If Google owns Youtube, then Google bought Youtube. Is that true or false?\\
A) True\\
B) False\\
Answer: B) False. Owning does not imply buying, the ownership may come from other means.\\
If John went to the mall, then John drove to the mall. Is that true or false?\\
A) True\\
B) False\\
Answer: B) False. John may have gone to the mall by other means.\\
If John drove to the mall, then John went to the mall. Is that true or false?\\
A) True\\
B) False\\
Answer: A) true. Driving is a means of going to the mall.\\
If John F. Kennedy was killed in Dallas, then John F. Kennedy died in Dallas. Is that true or false?\\
A) True\\
B) False\\
Answer:\\
\midrule \hline
\textbf{B. Few-shot Examples Instantiated Prompt for Attesting Hypothesis}\\ \hline
\midrule
Google bought Youtube. Is that true or false?\\
A) True\\
B) Unknown\\
C) False\\
Answer: A) True.\\
Yoshua Bengio likes oak trees. Is that true or false?\\
A) True\\
B) Unknown\\
C) False\\
Answer: B) Unknown.\\
The sun rises from the west. Is that true or false?\\
A) True\\
B) Unknown\\
C) False\\
Answer: C) False.\\
Answer:\\
        \midrule \hline
    \end{tabular}
    \caption{Example instantiated prompts in Few-shot settings, for the sample ``\textsc{premise}: [Google bought Youtube], \textsc{hypothesis}: [Google owns Youtube]''. The few-shot prompts in part B are used throughout the main experiments in this paper. We also present an example of the prompts we use for the hypothesis-only measure as described in \S\ref{sec:prompt template for attesting hypothesis appendix}.}
    \label{tab:example_instantiated_prompts}
\end{table*}

\begin{table*}[t]
    \centering
    \small
    \begin{tabular}{p{12cm}}
        \midrule \hline
\textbf{C. Few-shot Chain-of-Thought Prompts}\\ \hline
If Google bought YouTube, then Google owns YouTube. Is that true or false?\\
A) True\\
B) False\\
Explanation: \\
  1. Analyze Premise: the premise describe Google bought YouTube.\\ 
  2. Analyze Hypothesis: the hypothesis state that Google owns the YouTube.\\ 
  3. Reasoning: company A bought company B, it means that the company B belongs to company A now. So the premise entails hypothesis.\\
Answer: A) True.\\
If Google owns YouTube, then Google bought YouTube. Is that true or false?\\
A) True\\
B) False\\
Explanation:\\ 
  1. Analyze Premise: the premise state that Google owns the YouTube now.\\
  2. Analyze Hypothesis: the hypothesis describe Google bought YouTube.\\
  3. Reasoning: owning does not imply buying, the ownership may come from other means. So the premise does not entail hypothesis.\\
Answer: B) False.\\
If John went to the mall, then John drove to the mall. Is that true or false?\\
A) True\\
B) False
Explanation:\\ 
  1. Analyze Premise: the premise state that a person go to someplace.\\
  2. Analyze Hypothesis: the hypothesis describe a person went to someplace by driving car.\\
  3. Reasoning: A person may have gone to the place by other ways, so the premise does not entail hypothesis.\\
Answer: B) False.\\
If John drove to the mall, then John went to the mall. Is that true or false?\\
A) True\\
B) False\\
Explanation: \\
  1. Analyze Premise: the premise describe a person went to someplace by driving car.\\
  2. Analyze Hypothesis: the hypothesis state that a person go to someplace.\\
  3. Reasoning: Driving present the way to went to the place, so the premise entails hypothesis.\\
Answer: A) true.\\
If John F. Kennedy was killed in Dallas, then John F. Kennedy died in Dallas. Is that true or false?\\
A) True\\
B) False\\
Answer:\\
         \hline
    \end{tabular}
    \caption{Examples of CoT prompts, contains 3 steps: (1) analyze premise. (2) analyze hypothesis. (3) finding the relation between hypothesis and premise.}
    \label{tab:example_CoT_prompts}
\end{table*}

\end{document}